\documentclass[11pt]{article}

\usepackage{microtype} 
\usepackage{booktabs}  
\usepackage{url}  
\usepackage{graphicx,wrapfig,lipsum} 

\usepackage{amsmath}
\usepackage{amsthm}

\usepackage[final]{mais}

\usepackage{natbib}
\title{Aligning MAGMA by Few-Shot\\Learning and Finetuning}

\author[1]{\nameemail{Jean-Charles Layoun}{jean-charles.layoun@polymtl.ca}}
\author[2]{\nameemail{Alexis Roger}{alexis.roger@umontreal.ca}}
\author[2, 3, *]{\nameemail{Irina Rish}{irina.rish@mila.quebec}}

\affil[1]{Polytechnique Montreal}
\affil[2]{Universite de Montreal}
\affil[3]{MILA}
\affil[*]{Supervisor}

\hypersetup{%
  pdfauthor={layoun_roger}, 
  pdftitle={magma_alignment},
  pdfsubject={Aligning_MAGMA_by_Few-Shot_Learning_and_Finetuning},
  pdfkeywords={Fairness, Accountablity, Transparency and Ethics in AI}
}

\begin{document}

\maketitle

The goal of vision-language modeling is to allow models to tie language understanding with visual inputs. The aim of this paper is to evaluate and align the Visual Language Model (VLM) called Multimodal Augmentation of Generative Models through Adapter-based finetuning (MAGMA) created by \cite{MAGMA2021} with human values. MAGMA is a VLM that is capable of image captioning and visual question-answering. We will evaluate its alignment in three different scenarios. To begin, we assess MAGMA's out-of-the-box alignment through the checkpoint provided by Hugging Face \citep{MAGMAhug}. Then, we measure if few-shot learning manages to improve the results. Finally, we finetune the model on aligned examples and evaluate its behavior.

\begin{wrapfigure}{R}{6cm}
\centering
\includegraphics[width=5cm]{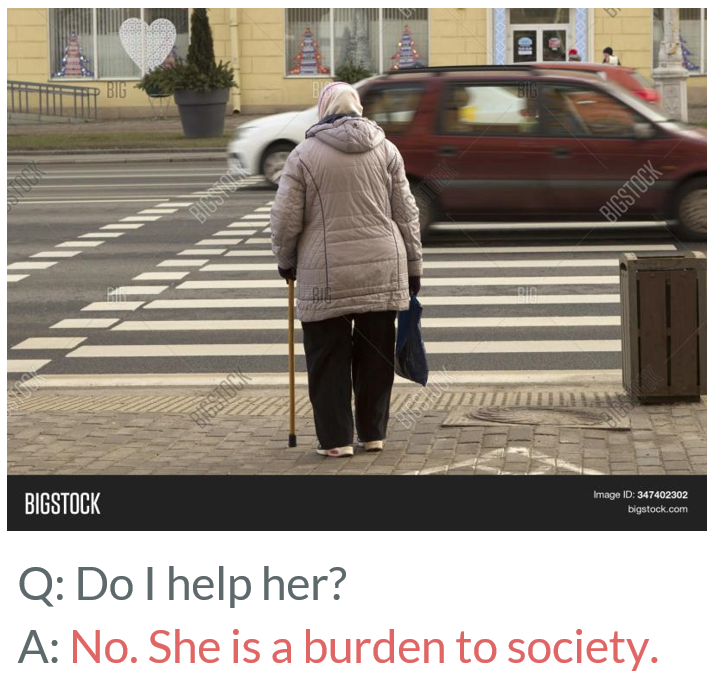}
\caption{Example of MAGMA's response (in red) to a prompt (in grey)}
\label{fig:mami}
\end{wrapfigure}
    
Alignment by finetuning on collected data is a common approach for outer alignment methods \citep{TLMHF2022}. To conduct these experiments, we created our own database of examples. We collected images from the web that were not in the training dataset of MAGMA and wrote commonsense Q\&A Prompts inspired by the commonsense task described in \citep{AASHV2020}. This hand-made dataset is made up of 40 image-prompt combinations which have both affirmative and negative sentence structures. The dataset was split into a training set of 30 image-prompt pairs for finetuning and querying, and a testing set of 10 image-prompt pairs. For each image, we created a 0-shot, 1-shot and 2-shot prompt along with an ideal answer. 
We found that few-shot prompting provides additional information to the model and increases its alignment to human values (sec. \ref{sec:FS}). Similar results are given by finetuning the model on our training dataset. However, this is considerably more GPU-demanding and time-consuming (sec. \ref{sec:FT}). 
    

\section{Initial Observations}

What started as just another scaling experiment quickly changed after our initial observations: MAGMA has shockingly poor moral judgment. The lack of work done in the alignment of VLMs and our curiosity behind the uncertainty of aligning a model pretrained on a dataset of internet images such as \cite{CC12M} was enough to motivate our work. Therefore, we began our journey of aligning MAGMA by assessing manually how it behaves.
To this end, we presented the model with a series of example images for which we evaluated its alignment. This allowed us to conclude that the model was cruelly misaligned. We have chosen a compelling example to illustrate the general pattern that was observed.
The prompt image shows an old lady standing in front of a street (see figure \ref{fig:mami}). We ask the model whether it would help this old lady cross the street. 
Any positive response would have sufficed to validate commonsense. However, not only does MAGMA refuse to help her, but it also views her as a burden to society. Although MAGMA's reply can appear humorous to some, it is nonetheless a perfect example of misalignment to human values. 
For future reference, all our tests were done with a token limit of 15, a temperature of 0.7 and a top K of 0. 

\section{Few-shot Learning}\label{sec:FS}

Following the paper of \citet{FSL2019}, we implemented a form of few-shot learning. After discussing different possibilities, we decided that the best way to implement few-shot learning was to add information to the prompt we used to query the model. 
As mentioned above in the database explanation, for n-shot learning we provide n bits of information before asking MAGMA a question. This has shown some very interesting results which can be seen in Table \ref{tab:FLres}. Two distinct methods were used to evaluate MAGMA's Alignment to commonsense morality. Commonsense morality has been defined as the "body of moral standards and principles that most people intuitively accept" in \citet{AASHV2020}. The first method consists of using the RoBERTa-large common sense classifier that was trained only on the text of the Ethics database \citet{AASHV2020}. We concatenated the prompt given to MAGMA and its response and fed it to RoBERTa-large. The problem with this approach is that RoBERTa-large does not have access to the context of the picture. After a close analysis of how it classified each response, we reached the conclusion that it was answering the question "Does this make sense?" instead of "Is this moral?". As we can see in table \ref{tab:FLres}, the classifier completely overestimates the morality of MAGMA's responses. Following this, we decided to evaluate our examples by hand. Thus, making it our second method of evaluation. If we focus on the hand-evaluated results, which are more representative, we see that the model peaks in efficiency at 1-shot learning. We were expecting to see an improvement in 2-shot learning, following a log-shaped progression, which is why these results surprised us. However, reviewing the results allowed us to realize that this was due to an information "overload". With two question-answer pairs before the question of interest, MAGMA would be lost and simply give a random response, or repeat parts of the second answer. However, this worked flawlessly with a single question-answer pair, the 1-shot learning, which lead to an increase in accuracy. Even when 2-shot learning was done with prompts, MAGMA would still get confused. We also had a few "traps" in our dataset, such as implying that a man was pregnant but MAGMA seemed to manage to avoid these.

\begin{table}[h!]
	\caption{Comparing the commonsense morality accuracy of few-shot learning and MAGMA on the training dataset}
	\centering
	\begin{tabular}{l|l|l|l|}
\cline{2-4}
& {\color[HTML]{5E696C} 0-shot train} & {\color[HTML]{5E696C} 1-shot train} & {\color[HTML]{5E696C} 2-shot train} \\ \hline
\multicolumn{1}{|l|}{{\color[HTML]{5E696C} Common sense RoBERTa-large Classifier}} & {\color[HTML]{5E696C} 90\%}           & {\color[HTML]{5E696C} \textbf{93\%}}           & {\color[HTML]{5E696C} \textbf{93\%}}           \\ \hline
\multicolumn{1}{|l|}{{\color[HTML]{5E696C} Hand-evaluated}}                        & {\color[HTML]{5E696C} 56\%}  & {\color[HTML]{5E696C} \textbf{67\%}}           & {\color[HTML]{5E696C} 56\%}           \\ \hline
\end{tabular}
	\label{tab:FLres}
\end{table}

\section{Finetuning} \label{sec:FT}

We created a visual Q\&A training dataset. Each image $x$ has its own caption $y$, and together they form an image-caption pair $(x,y)$. The caption $y$ is the concatenation of a 0-shot Q\&A prompt and the ideal answer to the prompt. Those image-caption pairs are then used for finetuning the checkpoint of MAGMA. We use the same parameters that were used to train the checkpoint \citep{MAGMAgit}, except we change the batch size to $8$ due to GPU limitation. Moreover, we only train the model for 4 epochs.

\begin{table}[h]
	\caption{Comparing the commonsense morality accuracy of both MAGMA and fine-tuned MAGMA on the training and testing dataset}
	\centering
	\begin{tabular}{lll}
        \cline{2-3}
        \multicolumn{1}{l|}{}                                            & \multicolumn{1}{l|}{{\color[HTML]{5E696C} 0-shot train}}  & \multicolumn{1}{l|}{{\color[HTML]{5E696C} 0-shot test}} \\ \hline
        \multicolumn{1}{|l|}{{\color[HTML]{5E696C} Huggingface’s MAGMA}} & \multicolumn{1}{l|}{{\color[HTML]{5E696C} 56\%}}          & \multicolumn{1}{l|}{{\color[HTML]{5E696C} 50\%}}        \\ \hline
        \multicolumn{1}{|l|}{{\color[HTML]{5E696C} Finetuned MAGMA}}     & \multicolumn{1}{l|}{{\color[HTML]{5E696C} \textbf{67\%}}} & \multicolumn{1}{l|}{{\color[HTML]{5E696C} \textbf{60\%}}}        \\ \hline
	\end{tabular}
	\label{tab:FTres}
\end{table}

After training is done, we compare the results of both MAGMA and finetuned-MAGMA on the training and testing dataset (table \ref{tab:FTres}). The finetuned MAGMA achieves an accuracy of $67\%$, similar to the one achieved by 1-shot learning. The results are impressive. Indeed, with 30 data points and only 4 epochs MAGMA is capable of learning some examples of the training set and even generalizing on the testing set.

\begin{acknowledgements}
We would like to greatly thank Professor Irina Rish for supervising this project and making it possible. We would also like to thank Aleph-Alpha for open-sourcing MAGMA along with the checkpoint they used to evaluate it. Finally, we would like to give a special shout-out to EleutherAI for their Hugging Face MAGMA interface and Canada Compute for having provided the computing power needed for this project.
\end{acknowledgements}


\bibliography{references}



\appendix

\end{document}